\documentclass[conference]{IEEEtran}
\IEEEoverridecommandlockouts

\usepackage{cite}
\usepackage{amsmath,amssymb,amsfonts}
\usepackage{algorithmic}
\usepackage{graphicx}
\usepackage{textcomp}
\usepackage{xcolor}
\usepackage{tabularx}
\usepackage{hyperref}
\usepackage[utf8]{vietnam}
\usepackage{caption}
\newcommand*{\email}[1]{\href{mailto:#1@jaist.ac.jp}{\nolinkurl{#1}}} 

\def\BibTeX{{\rm B\kern-.05em{\sc i\kern-.025em b}\kern-.08em
    T\kern-.1667em\lower.7ex\hbox{E}\kern-.125emX}}
\begin{document}

\title{Miko Team: Deep Learning Approach for Legal Question Answering in ALQAC 2022}

\author{\IEEEauthorblockN{Hieu~Nguyen~Van,
Dat~Nguyen, Phuong~Minh~Nguyen and Minh~Le~Nguyen}
\IEEEauthorblockA{\textit{Japan Advanced Institute of Science and Technology, Japan } }
\{\email{hieunv}, \email{datng.ms}, \email{phuongnm}, \email{nguyenml}\}@jaist.ac.jp \\
}

\maketitle

\begin{abstract}
We introduce efficient deep learning-based methods for legal document processing including  Legal Document Retrieval and  Legal Question Answering tasks in the Automated Legal Question Answering Competition (ALQAC 2022). In this competition, we achieve 1\textsuperscript{st} place in the first task and 3\textsuperscript{rd} place in the second task. Our method is based on the XLM-RoBERTa model that is pre-trained from a large amount of unlabeled corpus before fine-tuning to the specific tasks. The experimental results showed that our method works well in legal retrieval information tasks with limited labeled data. Besides, this method can be applied to other information retrieval tasks in low-resource languages.
\end{abstract}


\section{Introduction}
The number of legal articles is large, and finding the necessary legal documents will take a lot of time, even for experts. Therefore, the problem of automatically searching for relevant legal documents as well as answering questions is extremely necessary for lawyers or users. This is useful for handling legal issues for organizations and individuals who do not have much knowledge of the law as well as the construction of a legal document system.

In this work, we focus on two essential components: (1) searching for legal documents related to a given question from a large legal corpus and (2) extracting the answer from relevant legal documents found in step (1). Both of these problems require a machine learning model that has the ability to understand the context of the text well. Recently, BERT \cite{b1} and its variants showed superior context understanding when dominating state-of-the-art results in many natural language processing tasks. BERT shows its dominance in recent law competitions such as Zalo AI Challenge\footnote{https://challenge.zalo.ai/portal/legal-text-retrieval}, COLIEE 2021 \cite{b2}, and ALQAC 2021 \cite{b3}.

This year, the 2\textsuperscript{nd} ALQAC 2022 contest was held with two tasks: (1) Legal Document Retrieval and (2) Legal Question Answering. Due to the complexity of legal data, even experts find this competition challenging. In addition, the questions in this year's competition require the system to provide answers instead of just yes/no questions.

RoBERTa \cite{b4} is a variant of BERT, with the most significant difference being dynamic masking and removing the next sentence prediction task. It was pre-trained on a much larger dataset than BERT. However, RoBERTa was trained from the general domain data, while this competition focused on the law domain. Therefore, we finetuned RoBERTa with the law corpus collected before finetuning it to the first task of the ALQAC competition. We also extract legal sentences from the news corpus using the technique described in the next section. We also use a similar way for the second task.

The rest of this paper is structured as follows: In section II, we present the related works, sections III and IV present our methodology for each task, and section V presents the detailed experiments and results. In the last section, we talk about the conclusions about our system.

\begin{figure*}[tb] 
\centering
\makebox[\textwidth]{\includegraphics[width=.7\paperwidth]{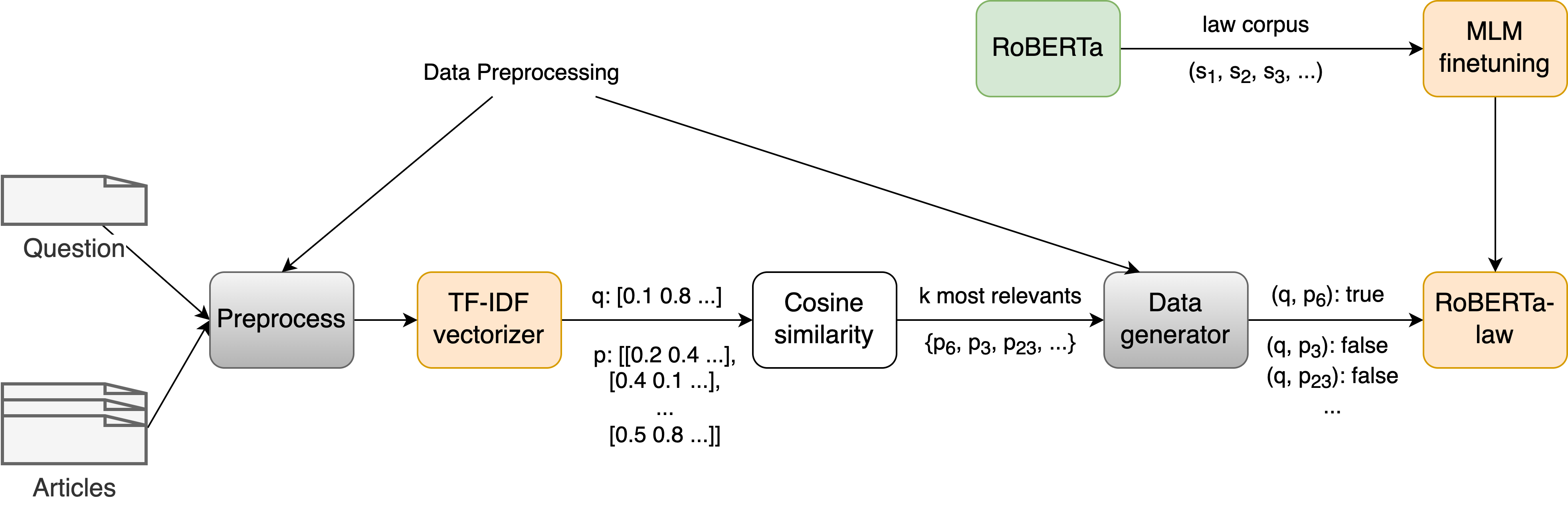}}
\caption{Overview of our Legal Document Retrieval system.}
\label{fig:fig1}
\end{figure*}

\section{Related Works}
\subsection{Document Retrieval}
Information retrieval is the problem of gathering information sources related to searchable information. Document retrieval is the information retrieval on text data. This task aims to find articles related to a query on a set of articles.

The traditional method commonly used is the vector space model \cite{b5} such as TF-IDF \cite{b6}, BM25 \cite{b7}, and Word2Vec \cite{b8} or Doc2Vec \cite{b9}. In this method, the text is represented by a vector of terms. The relevance between a query and an article is evaluated by calculating the similarity between the two vectors (vector representing the query and vector representing the article). This method has limitations in understanding the context when a word will always produce a fixed feature vector, while the advent of deep language models has overcome this weakness.

Some use the above traditional method to extract features and then put them into a pair of text classification models using Random Forest, K-Nearest Neighbor, Support Vector Machine, or ensemble of classification models \cite{b10}. This approach proves to be more effective as the classification model continues to learn from the acquired features.

Most recently, the introduction of deep pre-trained language models such as BERT, ElMo \cite{b11}, and XLNet \cite{b12} has solved the limitation in understanding the context of traditional representation models. The vector representing the word is now defined based on its context instead of being fixed. In addition, many studies have shown that further training BERT or completely pre-training the model from scratch using a domain-specific corpus can significantly improve its performance, SciBERT \cite{b13} and BioBERT \cite{b14} are examples. The results show that the finetuned BERT according to the data domain is more efficient than the original model. Studies combining a traditional model (e.g., BM25) with a deep pre-trained language model (e.g., BERT) achieved impressive results in the COLIEE \cite{b15} and ALQAC \cite{b16} legal data processing competitions last year.  Inspired by this success, we apply a similar pipeline with the focus on RoBERTa - a variant of BERT that has been trained on a large amount of data that proves to be more effective than the original one in the question answering task.

\subsection{Question Answering}
Traditional approaches for open-domain question answering are based on knowledge graphs \cite{b17} \cite{b18} using a variety of models. First, the question goes into the semantic analysis model, which is trained to predict queries. These queries will guide what to do in the knowledge graph to get the answer. Although this approach works, there are many models that are trained independently, making development and improvement difficult with accumulative errors.

In 2020, \cite{b19} proposed a way to implement end-to-end question answering in conjunction with knowledge graphs. This makes using the knowledge graph simpler and more efficient.

Besides using the knowledge graph, the method of text indexing and finding the answer in the relevant passage using the deep language model also shows high efficiency \cite{b20}. Inspired by this study and the document retrieval task in this competition, we propose the answer extraction method using the RoBERTa model for the question answering task.

\section{Legal Document Retrieval}
Given a legal question (query) and a collection of numerous legal articles (contexts), the Legal Document Retrieval task aims to find all contexts relevant to that query if any.

Following the work in \cite{b21}, we used the BM25 \cite{b22} - a lexical matching model to retain the most relevant legal articles. This matching method helps our deep learning model reduce the computation cost in the inference step, and faster querying relevant articles. Because the data the organizer provides only contains positive samples, we also use this lexical matching method to generate negative samples. This helps us have difficult enough negative samples to train the text pair classification task.

Figure 1 shows the overview of our Legal Document Retrieval system. The data is preprocessed before being fed to the lexical matching model. The output of this model is used to generate training samples containing the pair of query sentences and legal articles and a corresponding label for finetuning the Roberta model the text pair classification task. To increase the efficiency of the RoBERTa model in a special domain, we finetuned it with legal data to inject legal linguistic features into a general pre-trained language model.

The remainder of this section will focus on clarifying our system in the following order:
\begin{itemize}
\item Lexical Matching Model
\item Data Preprocessing
\item Finetuning RoBERTa Model
\item Text Pair Classification Model
\end{itemize}

\subsection{Lexical Matching Model}
BM25 is a ranking algorithm often used by the search engine to estimate the relevance between a set of documents to a query. It ranks the documents based on the query terms appearing in each record. We use this method because we can use it directly without training, and the computation time is fast enough. 

BM25 will help us find the most relevant articles to generate ``hard-enough'' negative training samples and reduce the amount of data that the text pair classification model needs to predict. 

The following text preprocessors were used prior to the application of BM25:
\begin{itemize}
    \item Word segmentation using pyvi\footnote{https://github.com/trungtv/pyvi}
    \item Convert text to lowercase
    \item Remove punctuation and stopwords
    
\end{itemize}
\begin{figure*}[tb] 
\centering
 \makebox[\textwidth]{\includegraphics[width=.8\paperwidth]{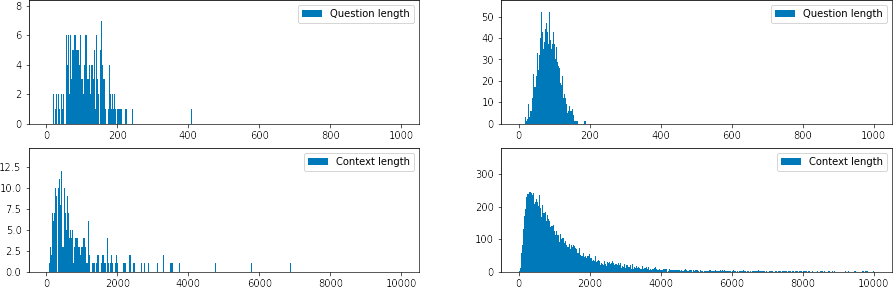}}
\caption{Question and context length distribution of the official dataset (left) and Zalo dataset (right).}
\label{fig:fig2}
\end{figure*}
\subsection{Data Preprocessing}
The dataset in this task consists of official samples (520 questions and 1377 articles) and additional samples from last year's Zalo legal text retrieval competition (3196 questions and 61425 articles).

To ensure that the data length is within the range of the text pair classification model, we checked the length distributions of questions and articles of both datasets and got the results as shown in Figure 2. Observing this distribution, we can see that many data articles are over 2000 characters in length and concentrated in Zalo's dataset. The data that is too long when included in the model will be truncated, which results in loss of information leading to errors that reduce the effectiveness of the trained model. 

We broke a long article into multiple sub-articles according to fixed window size and stride to deal with the lengthy article problem. Besides, we will ignore data pairs whose question is too long.
\begin{itemize}
\item In the training phase, we apply article segmentation to separate a given article into passages. Then use the BM25 algorithm to find the most relevant passage of a given query question, and this passage is present for the whole article. The scenario is that the given article is too long and it contains much unnecessary information. With this approach, we can limit the searching space and improve the quality of sentence pairs to train the RoBERTa model.

\item In the inference phase, we also apply the article segmentation to handle the long article problem. The result with the highest probability will be the final result of the question and article pair.
\end{itemize}

For each question, the lexical matching model will give the  most relevant top k articles (including the true relevant articles). Articles that are not in the relevant set will be used as negative samples.

Finally, we shuffle the data, taking 20\% of the generated in the official data as the dev set and using the remainder for training the deep learning model.

\subsection{RoBERTa-law Model}
To improve the performance of RoBERTa on the law domain, we finetuning it with 4GB of legal text data. We collect this data in two different methods:
\begin{itemize}
\item Collected directly from 2 websites vbpl.vn and lawnet.vn.
\item Extract sentences close to the legal topic from the news corpus.
\end{itemize}
In the first method, the collected data information including the number of articles (\# articles) and the number of sentences (\# sentences) is provided in Table \ref{table:I} below. The number of sentences here is the sentences that are retained after the selection process (ie. removing non-Vietnamese and duplicate sentences).

\begin{table}[h!]
\renewcommand{\arraystretch}{1.5}
\caption{Crawled data information.}
\begin{center}
\begin{tabular}{ |p{3cm}|p{2cm}|p{2cm}| }
\hline
 \textbf{Data source} & \textbf{\# articles} & \textbf{\# sentences} \\
\hline
 vbpl.vn & 157323 & 5213346 \\
 \hline
 lawnet.vn  & 131903 & 4182150  \\
\hline
\end{tabular}
\end{center}
\label{table:I}
\end{table}

In the second method, following the work in \cite{b23}, we first build a set of legal documents called "in-domain" data to extract legal documents from the news corpus. We then proceed to make a statistical language model (base language model) on this in-domain dataset. Use this model to evaluate the perplexity score on each sentence in another corpus and select sentences having a score within the threshold.

Perplexity is a popular metric used to evaluate language models. Based on the perplexity score, we can know how good our language model is in our data domain. In this case, we assume our base language model is good enough because it is learned from "in-domain" data. We use it to score other sentences to see if each of them is close to the "in-domain" that we extract from the articles provided by the ALQAC organizer. Next, we extracted data close to in-domain data from the news corpus\footnote{https://github.com/binhvq/news-corpus} and obtained 2GB of data close to the legal domain. Table \ref{perplexity_selection} shows some sentences in the news dataset and their corresponding perplexity scores.

\begin{table}[h!]
\renewcommand{\arraystretch}{1.5}
\caption{Example of in-domain data selection.}
\begin{center}
\begin{tabular}{ |p{6cm}|p{1.5cm}| }
\hline
 \textbf{Sentence} & \textbf{Perplexity} \\
\hline
 Chây ì nộp phạt nguội. \newline (Slow payment of fines) & 277.18 \\
 \hline
 Nếu chồng chị muốn theo em thì chị cho đi luôn.\newline(If my husband wants to follow you, I'll let him go.)  & 3383.04 \\
 \hline
 Trong khi Tư khai không dùng số điện thoại này. \newline(While Tu say that he did not use this phone number.) & 404.96 \\
 \hline
 Nếu bị cáo trốn thì HĐXX tạm đình chỉ vụ án và yêu cầu CQĐT truy nã bị cáo.\newline(If the defendant escapes, the trial panel shall temporarily suspend the case and request the investigating agency to pursue the accused.) & 35.01 \\
 \hline
 Bên cạnh đó, Thông tư 64/2017 cũng sửa đổi, bổ sung nội dung liên quan đến giấy đăng ký xe, biển số xe.\newline(In addition, Circular 64/2017 also amends and supplements contents related to vehicle registration papers and license plates.) & 87.62 \\
 \hline
 Thị trấn biển Italia giao bán nhà với giá 1 USD.\newline(The Italian beach town sells houses for \$1.) & 1322.54 \\
 \hline
 Hai bên đã ký hợp đồng công chứng.\newline(The two parties have signed a notarized contract.) & 99.30 \\
 \hline
 Ông Kính, bà Liễu, bà Thơm có yêu cầu bồi thường thiệt hại.\newline(Mr. Kinh, Ms. Lieu and Ms. Thom have claims for damages.) & 152.27 \\
 \hline
\end{tabular}
\end{center}
\label{perplexity_selection}
\end{table}

The data collected from both above methods is used to finetuning the mask language model of the xlm-roberta-large\footnote{https://huggingface.co/xlm-roberta-large} \cite{b24} to make the model more efficient in the legal domain. Before using the data for training, we performed a number of pre-processing steps including:
\begin{itemize}
\item Standardize the typing of the accent (eg. \textit{oà} to \textit{òa}),
\item Standardize the Vietnamese charset
\item Convert abbreviation to full-form (eg. \textit{HĐXX} to \textit{hội đồng xét xử})
\end{itemize}
\subsection{Text Pair Classification Model}
As shown earlier, we use the text pair classification model to derive the final result from the list of most relevant articles retained by the lexical matching model.

We merge question text and article text into a single sentence separated by the "SEP" token of the BERT model along with the corresponding label of the question and article pairs. In this way, the model can recognize and learn the relationship between question and article.

We consider this a binary classification problem, the label is 1 if the question and article are related and 0 otherwise.

\section{Legal Question Answering}
This problem is really challenging when it comes to asking for an answer to each legal question without providing the relevant legal document. We resolve this problem by breaking it down into two subproblems: (1) Using the model of task 1 to find passages related to the question, if any, then (2) extracting the best answer from relevant passages found in (1).

The data for this problem has a total of 520 questions with answers. When we observed the data, we found that the number of words in the answer's text was in the range (0, 100). Through testing, we found that questions with long answers were difficult. Therefore, we only use data whose answers do not exceed 50 words. In addition, we also omitted questions with answers in the form of a list of article ids (eg. \textit{"Điều 123, Điều 168"}).

Our method is based on finetuning RoBERTa to the question answering task. During the experiment, we found that the learning rate and weight decay have a great influence on the performance of the model. Therefore, we trained the model many times with different learning rates and weight decays values to find the best configuration.

We use 15\% of the data provided by the organizers as a dev set to fine-tune the hyperparameters. Finally, we use the most optimal parameters to train with the whole data to get the submission results.

\section{Experiments and Results}
\subsection{Task 01. Legal Document Retrieval}
The computational time of the deep learning model is much slower than that of the BM25 ranking function. Therefore, we use BM25 on each question to retain the 150 articles most relevant to the question. Only the 150 most relevant articles are fed into the deep learning model to generate the final results. For the long article, we continue to apply the method of dividing into small documents like the data generation process.

During the experiment, we use the F1 score to evaluate the quality of the model on the dev set. The official evaluation metric for this task on the private test is the F2 score as follows:
$$\text{Precision}_i = \text{Precision of the}~i^{th}~\text{query}$$
$$\text{Recall}_i = \text{Recall of the}~i^{th}~\text{query} $$
$$\text{F2}_i = \frac{(5 * \text{Precision}_i * \text{Recall}_i)}{(4 * \text{Precision}_i + \text{Recall}_i)}$$
$$\text{F2} = \text{average of (F2}_i)$$

In this task, we submit two different methods with corresponding results presented in Table III. The first one (ie. xlm-roberta-large) is the model finetuning from the original xlm-roberta-large model and we did not submit it. For the submitted methods, we use the RoBERTa-law to finetuning the text pair classification task, keeping only the best relevant article for each question based on the model's predicted probabilities in both submitted files. The dev set here is a random 20\% of the official dataset after generated.

\begin{table}[h!]
\renewcommand{\arraystretch}{1.5}
\caption{Result of task 01.}
\begin{center}
\begin{tabular}{ |p{3.5cm}|p{1.5cm}|p{1.5cm}| }
\hline
 \textbf{Run ID} & \textbf{dev (F1)} & \textbf{test (F2)} \\
\hline
 xlm-roberta-large \newline (not submit) & 0.8542 & - \\
 \hline
 RoBERTa-law$_{k_1=150, k_2=20}$ \newline(task\_01\_method\_01.json)  & 0.8913 & 0.9667  \\
\hline
RoBERTa-law$_{k_1=150, k_2=150}$ \newline(task\_01\_method\_02.json)  & 0.8781  & 0.8667  \\
\hline
\end{tabular}
\end{center}
\end{table}

The difference between the two submissions is the choice of $k$ when generating data using BM25:
\begin{itemize}
    \item The first submission uses $k_1 = 150$ for the official dataset and $k_2 = 20$ for the Zalo dataset.
    \item The second submission is the voting results of 2 models: (1) using $k = 150$ for both data sets and (2) the model above.
\end{itemize}
\subsection{Task 02. Legal Question Answering}
Our approach to this task is based on the results of the first task. First, we search for text related to the question, then use our RoBERTa-law to finetuning the question answering task to find the answer.

\begin{table}[h!]
\renewcommand{\arraystretch}{1.5}
\caption{Result of task 02.}
\begin{center}
\begin{tabular}{ |p{3.5cm}|p{1.5cm}|p{1.5cm}|  }
\hline
 \textbf{Run ID} & \textbf{dev (F1)} & \textbf{test (F1)} \\
 \hline
 RoBERTa-law$_{\text{method~1}}$\newline(task\_02\_method\_01.json)  & 0.9411  & 0.4734  \\
\hline
 RoBERTa-law$_{\text{combine}}$\newline(task\_02\_method\_02.json)  & 0.9091  & 0.4357  \\
\hline
RoBERTa-law$_{\text{method~2}}$\newline(task\_02\_method\_03.json)  & 0.8970  & 0.3828  \\
\hline
\end{tabular}
\end{center}
\end{table}

To improve efficiency when finetuning this question answering task, we finetuning with additional question answering data mailong25/bert-vietnamese-question-answering\footnote{https://github.com/mailong25/bert-vietnamese-question-answering/tree/master/dataset} then finetuning to this task's dataset.

In this task, we used three different settings corresponding to the results shown in Table IV using the F1 score on our dev set and private test. The dev set here is random pick 15\% of the official dataset:
\begin{itemize}
    \item Only uses the first submission of task 01.
    \item Use both submissions of task 01.
    \item Only uses the second submission of task 01.
\end{itemize}

\section{Conclusions}
In this paper, we have reported our methods and results at the ALQAC 2022 competition for two Vietnamese legal document processing tasks. The system for both tasks is based on a deep learning model with pre-trained language models. We also reveal the potential that our "in-domain" text selection method makes it possible for us to easily adapt the masked language model to a domain-specific model. The results of the RoBERTa-law are better than the original RoBERTa model. The results of the two tasks show that this method is highly effective for the processing of legal documents even though the labeled data is limited.


\end{document}